\documentclass[a4paper, 10pt, conference]{ieeeconf}
\IEEEoverridecommandlockouts
\usepackage{cite}
\usepackage{amsmath,amssymb,amsfonts}
\usepackage{algorithmic}
\usepackage[]{algorithm2e}
\usepackage{graphicx}
\usepackage{textcomp}
\usepackage{xcolor}
\usepackage{subcaption}
\usepackage{float}
\usepackage{hyperref}
\usepackage{titlesec}
\usepackage{listings}
\usepackage{multirow}
\usepackage{amssymb}
\usepackage{pifont}
\usepackage{lscape}
\usepackage{array}
\usepackage{float}
\newfloat{footnote}{hb}

\newcolumntype{P}[1]{>{\centering\arraybackslash}p{#1}}

\begin{document}
\title{\LARGE \bf Multi-label Annotation for Visual Multi-Task Learning Models}

\author{Gaurang Sharma$^{1}$, Alexandre Angleraud$^{1}$ and Roel Pieters$^{1}$
\thanks{$^{1}$Unit of Automation Technology and Mechanical Engineering, Tampere University, Tampere, Finland; {\tt\small firstname.surname@tuni.fi}}%
}

\IEEEoverridecommandlockouts 

\maketitle
\thispagestyle{empty}
\pagestyle{empty}

\begin{abstract}

Deep learning requires large amounts of data, and a well-defined pipeline for labeling and augmentation. Current solutions support numerous computer vision tasks with dedicated annotation types and formats, such as bounding boxes, polygons, and key points. These annotations can be combined into a single data format to benefit approaches such as multi-task models. However, to our knowledge, no available labeling tool supports the export functionality for a combined benchmark format, and no augmentation library supports transformations for the combination of all. 
In this work, these functionalities are presented, with visual data annotation and augmentation to train a multi-task model (object detection, segmentation, and key point extraction). The tools are demonstrated in two robot perception use cases.
\end{abstract}

\section{Introduction}\label{Introduction}
Advancements in deep learning have helped to address numerous problems in different domains (e.g., robotics\cite{Károly}, medicine\cite{PICCIALLI2021111} and agriculture\cite{kamilaris2018deep}). 
Perception in particular has  been of major focus, where single purpose models have been optimized and fine-tuned for one specific task at hand \cite{AlexNet}.  
More recently also multi-task models have been developed \cite{Caruana97, Vandenhende2022}, where a shared model concurrently learns multiple tasks. A fundamental aspect of these models is to increase data efficiency, reduce overfitting, and enable quick learning due to the use of auxiliary information. They use a unified single backbone structure having multiple descriptive heads to address technical computer vision challenges like boundary detection, semantic segmentation, object detection, etc. 
To effectively utilize multi-task models, multi-label annotations should be supported by the framework. This implies that images and the objects inside them can be annotated with multiple different labels (e.g., bounding boxes, polygons and key points), to be used for learning different tasks and thus generate multiple outputs. Compared to single-task models this is less computationally expensive and requires less annotated records. In addition, multiple outputs per image or object provides additional information compared to single-task models. 
\begin{figure}[t]
          \centering
          \includegraphics[width=\linewidth]{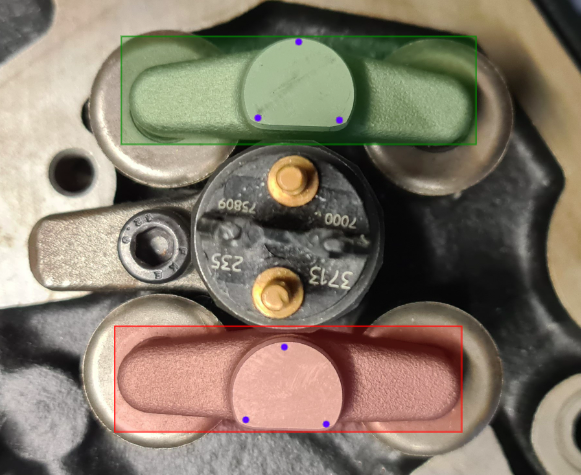}
        \caption{Motivation for multi-task detection models. Single-task models can detect an object (bounding box), but cannot verify whether their orientation is correct. Multi-task models can provide additional information from different model outputs. In this case key points (three blue points) indicate a correct (green box) and incorrect (red box) object placement.
        }\label{fig:motivation}
    \end{figure}
Fig. ~\ref{fig:motivation} depicts this visually in an object detection problem. Two objects are annotated and correctly detected, however, in this context, the lower object is incorrectly placed. From detection or segmentation alone, this cannot be determined, yet with the estimated key points as additional information (i.e., three blue points), incorrect placement can be detected.
Existing labeling and annotation solutions\cite{Pande2022} do not provide the combined annotation export functionality, which results in solutions that combine several export formats to accomplish a common task.
This served as motivation for the development of a novel data pipeline to provide domain-specific annotation and augmentation strategies. Best-of-breed libraries such as Albumentation \cite{Albumentations} and Label Studio \cite{LabelStudio} are integrated to create a generic data generation solution for training deep learning models. 
Our developments are evaluated by applications in object detection, object configuration estimation and depth estimation. 
In this work, our contributions are:
\begin{enumerate}
\item A novel pipeline that enables labeling of image datasets with different annotations and formats
\item The augmentation of this dataset without conversion to different formats
\item The validation of results in two industrial use cases with different computer vision tasks 
\end{enumerate}

\section{RELATED WORK}\label{related-work}

\begin{table*}[]
\centering
\caption{Well-known annotation tools and their main exported formats and supported annotation types. None of the existing tools support a combined annotation strategy in a benchmark format. }
\label{tab:AnnotationTools}
\begin{tabular}{|c|c|cccc|l|}
\hline
\textbf{} & \textbf{Export format} & \multicolumn{4}{c|}{\textbf{Annotation type}} & \multicolumn{1}{c|}{\textbf{Comment}} \\ \hline
\textbf{Annotation tool} & \multicolumn{1}{l|}{} & \multicolumn{1}{c|}{Bounding box} & \multicolumn{1}{c|}{Polygon} & \multicolumn{1}{c|}{Masks} & keypoints &  \\ \hline
\multirow{5}{*}{\textbf{Label Studio}\cite{LabelStudio}} & COCO \cite{COCO} & \multicolumn{1}{c|}{\checkmark} & \multicolumn{1}{c|}{\checkmark} & \multicolumn{1}{c|}{} &  &  \\ \cline{2-7} 
 & YOLO \cite{YOLOv7} & \multicolumn{1}{c|}{\checkmark} & \multicolumn{1}{c|}{} & \multicolumn{1}{c|}{} &  &  \\ \cline{2-7} 
 & PASCAL VOC XML \cite{pascal} & \multicolumn{1}{c|}{\checkmark} & \multicolumn{1}{c|}{} & \multicolumn{1}{c|}{} &  &  \\ \cline{2-7} 
 & JSON & \multicolumn{1}{c|}{\checkmark} & \multicolumn{1}{c|}{\checkmark} & \multicolumn{1}{c|}{\checkmark} & \checkmark & JSON supports any annotation type \\ \cline{2-7} 
 & NumPy arrays & \multicolumn{1}{c|}{} & \multicolumn{1}{c|}{} & \multicolumn{1}{c|}{\checkmark} &  &  \\ \hline
\multirow{5}{*}{\textbf{V7} \cite{V7}} & COCO \cite{COCO} \cite{COCO} & \multicolumn{1}{c|}{\checkmark} & \multicolumn{1}{c|}{\checkmark} & \multicolumn{1}{c|}{} &  &  \\ \cline{2-7} 
 & YOLO \cite{YOLOv7} & \multicolumn{1}{c|}{\checkmark} & \multicolumn{1}{c|}{} & \multicolumn{1}{c|}{} &  &  \\ \cline{2-7} 
 & PASCAL VOC XML \cite{pascal}& \multicolumn{1}{c|}{\checkmark} & \multicolumn{1}{c|}{} & \multicolumn{1}{c|}{} &  &  \\ \cline{2-7} 
 & Darwin JSON & \multicolumn{1}{c|}{\checkmark} & \multicolumn{1}{c|}{\checkmark} & \multicolumn{1}{c|}{\checkmark} & \checkmark & JSON supports any annotation type \\ \cline{2-7} 
 & PNG & \multicolumn{1}{c|}{} & \multicolumn{1}{c|}{} & \multicolumn{1}{c|}{\checkmark} &  &  \\ \hline
\multirow{5}{*}{\textbf{CVAT} \cite{CVAT}} & COCO object detection \cite{COCO} & \multicolumn{1}{c|}{\checkmark} & \multicolumn{1}{c|}{\checkmark} & \multicolumn{1}{c|}{} &  &  \\ \cline{2-7} 
 & COCO keypoint \cite{COCO} & \multicolumn{1}{c|}{} & \multicolumn{1}{c|}{} & \multicolumn{1}{c|}{} & \checkmark &  \\ \cline{2-7} 
 & YOLO \cite{YOLOv7} & \multicolumn{1}{c|}{\checkmark} & \multicolumn{1}{c|}{} & \multicolumn{1}{c|}{} &  &  \\ \cline{2-7} 
 & PASCAL VOC XML \cite{pascal} & \multicolumn{1}{c|}{\checkmark} & \multicolumn{1}{c|}{} & \multicolumn{1}{c|}{\checkmark} & \checkmark &  \\ \cline{2-7} 
 & MOTS \cite{Voigtlaender19CVPR_MOTS} & \multicolumn{1}{c|}{} & \multicolumn{1}{c|}{\checkmark} & \multicolumn{1}{c|}{} &  &  \\ \hline
\textbf{LabelBox} \cite{labelbox}& JSON & \multicolumn{1}{c|}{\checkmark} & \multicolumn{1}{c|}{\checkmark} & \multicolumn{1}{c|}{\checkmark} & \checkmark & Separate JSON for each annotation \\ \hline
{\textbf{LabelMe}\cite{labelme}} & JSON & \multicolumn{1}{c|}{\checkmark} & \multicolumn{1}{c|}{\checkmark} & \multicolumn{1}{c|}{\checkmark} & \checkmark &  JSON supports any annotation type\\ \hline
\textbf{Our work} & COCO \cite{COCO} & \multicolumn{1}{c|}{\checkmark} & \multicolumn{1}{c|}{\checkmark} & \multicolumn{1}{c|}{\checkmark} & \checkmark & \multicolumn{1}{c|}{} \\ \hline
\end{tabular}
\end{table*}

\subsection{Multi-task Learning}
Multi-task learning \cite{Caruana97} in the field of computer vision \cite{Vandenhende2022} has shown recent success with methods such as Cross-Stitch \cite{misra2016cross} and UberNet \cite{UberNet}. As comparison, single-task models are more expensive to train and require a larger amount of annotated data than multi-task models. Moreover, knowing which tasks should be trained together could aid in improving prediction quality and acting as a preventative measure against the occurrence of negative transfer \cite{NEURIPS2021_e77910eb}. Performance improvement of a specific task can thus lead to performance degradation of other tasks \cite{9506618}.
Currently, libraries such as YOLOR\cite{YOLOR} and YOLOv7\cite{YOLOv7} provide configuration features to facilitate multi-task learning.



\subsection{Image Annotation and Storing Formats}
Labeling data is a critical component for deep learning as it provides the ground truth for estimating errors. Incorrectly labeled datasets lead to poor training and higher validation losses \cite{Bhagat20181}. Nowadays, there are several machine learning based tools available that speed up the automatic labeling process and reduce human labor and time \cite{Sager2021}.
MS COCO\cite{COCO}, PASCAL VOC\cite{pascal}, and YOLO are some commonly used benchmarks with different storage format styles. MS COCO is a rich (.json) format that offers a large variety of annotation storing structures.
PASCAL VOC, on the other hand, supports a (.xml) structure that is useful for detection and segmentation tasks but has limitations in supporting key points. YOLO models accept (.txt) file format for detection tasks, however, YOLOv7\cite{YOLOv7} accepts polygon segmentation and key point representations as well.
While numerous labeling tools are available, each provides support for different export formats and annotation types (see Table \ref{tab:AnnotationTools}). For example, only a few tools offer direct labeling options for polygon segmentation, bounding boxes and key points simultaneously. And when the opportunity is provided, only a combined JSON export format is available, thereby not directly supporting state-of-the-art deep learning libraries. As a result, custom data loaders need to be developed to enable library-specific data formats. 


    \subsection{Image Augmentation}
        Data augmentation is a form of data expansion that can improve model performance by assisting in its generalization, robustness and convergence.         
        While augmentation is a well-known approach, libraries such as TensorFlow \cite{tensorflow2015-whitepaper} and PyTorch \cite{PyTorch} provide very few transformation strategies. Albumentations \cite{Albumentations}, on the other hand provides around six different libraries to apply a broad range of transformations within a single tool. However, if custom and complex transformations are required, then libraries like OpenCV \cite{opencv_library} can be utilized for basic image manipulation algorithms. 
        Smart augmentation techniques also exist \cite{SmartAugmentationLearning}, in which suitable augmentations are learned during the process of training a deep neural network. 
        

         \subsection{Object detection and Pose estimation}
         Traditionally, keypoint estimation, segmentation, and object detection were considered separate problems. The solution for object detection evolved in the last 20 years from hand-crafted features to deformable transformers \cite{ObjectDetectionIn20Years}. Research on segmentation started with thresholding and later led to methods such as semantic, instance, and panoptic segmentation. On the other hand, local features, called keypoints, became popular through corner and edge detectors \cite{COMBINEDCORNERANDEDGEDETECTOR}, and today there are several ways to predict them, including Faster R-CNN \cite{LocalkeypointBasedFasterR-CNN}, which is based on local keypoints.
         In the field of robotics and computer vision, segmentation masks are used to predict keypoints, as presented in  \cite{semantic_segmentation_keypoint_detection}. Recently, combinations of both to predict human and object poses is addressed by \cite{kPAM}. In this, COCO is used as the primary format, emphasizing its relevance.

 \section{Methodology}\label{Methodology}
     

In this section, we present our methodology which enables the annotation of images with multiple labels in multiple formats and the export of these in a generic format, for training a multi-task detection model. 

         \begin{figure*}[t]
          \centering
          \includegraphics[width=0.9\linewidth]{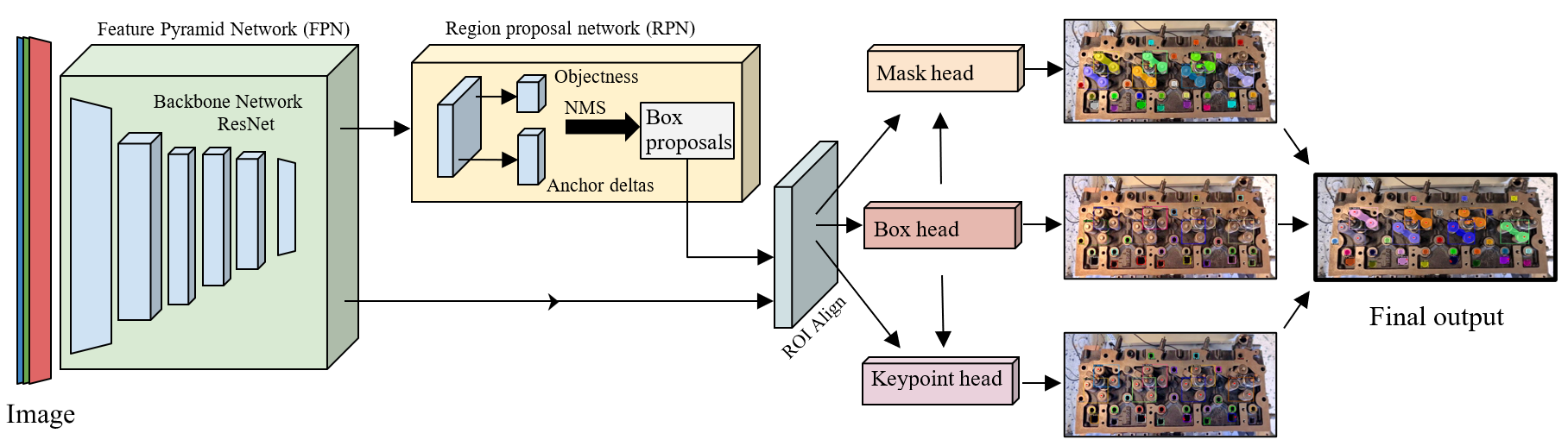}
        \caption{The multi-task learning architecture from Detectron2 \cite{Detectron2} supports the combined data annotations (bounding box, polygon and key points) for multi-task learning. The final output is the combination all three detection formats.}
        
        \label{fig:modelFramework}
        \end{figure*} 

\subsection{Architecture}
  The multi-task architecture of ResNet-50 FPN Keypoint R-CNN \cite{Detectron2} supports the combined data annotations and is utilized for training the detection model (see Fig. \ref{fig:modelFramework}). The output of the model contains all three detection formats, for object recognition, segmentation and feature extraction at the same time.
    
    \subsection{Data Annotation}
Our pipeline starts with the image annotation of polygons, key points and bounding boxes in Label Studio. As Label Studio does not provide the functionality for simultaneously annotating features for polygons and key points, our approach updates polygon configurations by adding a key points configuration script to the user interface, maintaining a single image via the 'toName' attribute. Bounding boxes are extracted from the polygon annotations by selecting the outer bounds of the polygon coordinates. Following, all annotations are exported into a JSON-min format and converted into COCO format, containing all three data annotation types in a single output format. 

    \subsection{Data Augmentation}

     Tools such as Albumentations \cite{Albumentations} offer a wide range of image transformations. 
     However, it does not offer augmentation support for polygon and run-length encoding COCO formats, limiting the support for an integrated augmentation strategy. 
    To enable the augmentation of multi-label annotated data, such as bounding boxes, key points and COCO segmentation formats, the following steps are performed.
     First, the combined data (in COCO format) is loaded and extracted into separate arrays, such as bounding boxes, key points, area, etc. 
     Second, polygons are converted to key points and appended to a key point array. 
     Third, the conversion of (x,y) key points to (x,y,v) is done to maintain COCO standards, followed by appending invisible key points to keep the number of labeled key points the same. 
    Dataset augmentation is then performed using Albumentation's key point transformation strategy, after which the key points are converted back to polygons. 
     As a result, this procedure enables the transformation of COCO polygon format by the Albumentation library.
        
    
    

\section{Results and Discussion}\label{ResultsAndDiscussion}
This section presents the results of our multi-label annotation pipeline, its detection results and a discussion on its limitations.

\subsection{Practical Use Cases and System Integration}
The use cases aim to detect different objects for automation purposes. Two assembly sets are selected including a diesel engine and a planetary gearbox, and their internal parts (e.g., rocker arms, pushrods, bolts, gears, housing, etc.). 
The diesel engine \cite{Sharma2023} and planetary gearbox \cite{Sharma2023_gear} datasets contain 195 and 150 RGB images, respectively, manually annotated by drawing polygons and labeling key points (see Fig. \ref{fig:diesel_annotation} and \ref{fig:planetary_annotation}). 
Augmentation is then performed on the annotated data by different transformation strategies, completing datasets of 280,000 and 170,000 images. 
Training of the multi-task models followed typical training and evaluation steps \cite{Detectron2} with hyper-parameters such as momentum, weight decay, and regularization to achieve loss convergence in about 2000 epochs.
The images were captured with an Intel Realsense D435 camera. All tools are open-source available from the dataset documentations \cite{Sharma2023, Sharma2023_gear}.

    \subsection{Object and Key point Detection}\label{Applications}
\begin{figure*}[htp]
\subcaptionbox{\label{fig:diesel_annotation}}{%
  \includegraphics[height=0.245\linewidth]{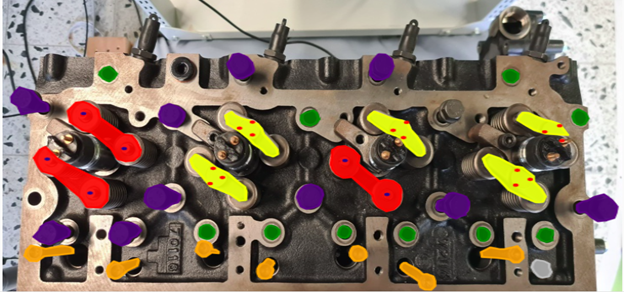} 
}\hfill
\subcaptionbox{\label{fig:diesel_detection}}{%
  \includegraphics[height=0.245\linewidth]{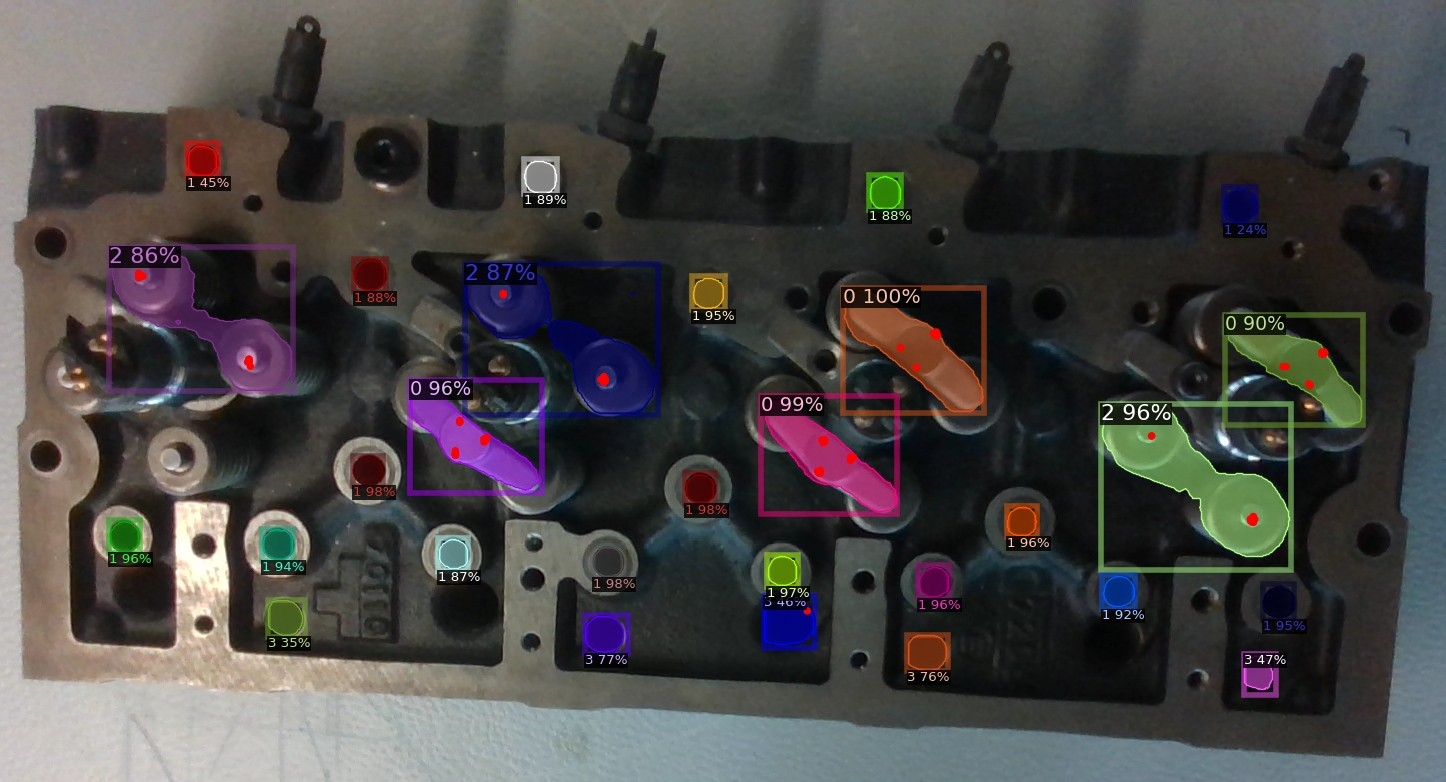} 
}   


\subcaptionbox{\label{fig:planetary_annotation}}{%
  \includegraphics[height=0.29\linewidth]{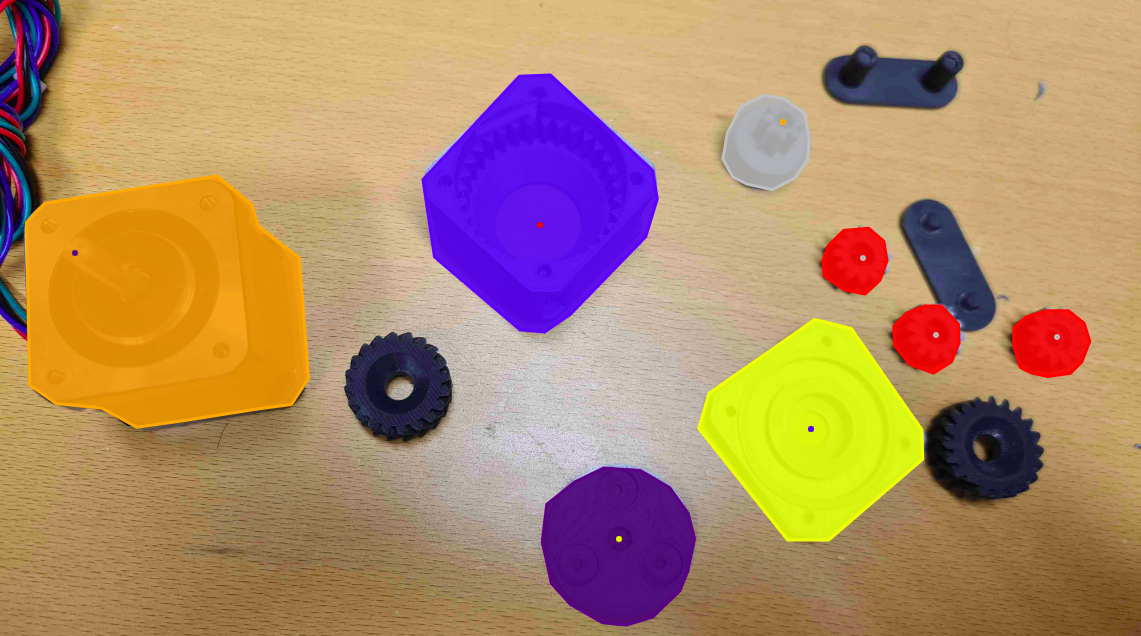} 
}\hfill 
\subcaptionbox{\label{fig:plan_detections}}{%
  \includegraphics[height=0.29\linewidth]{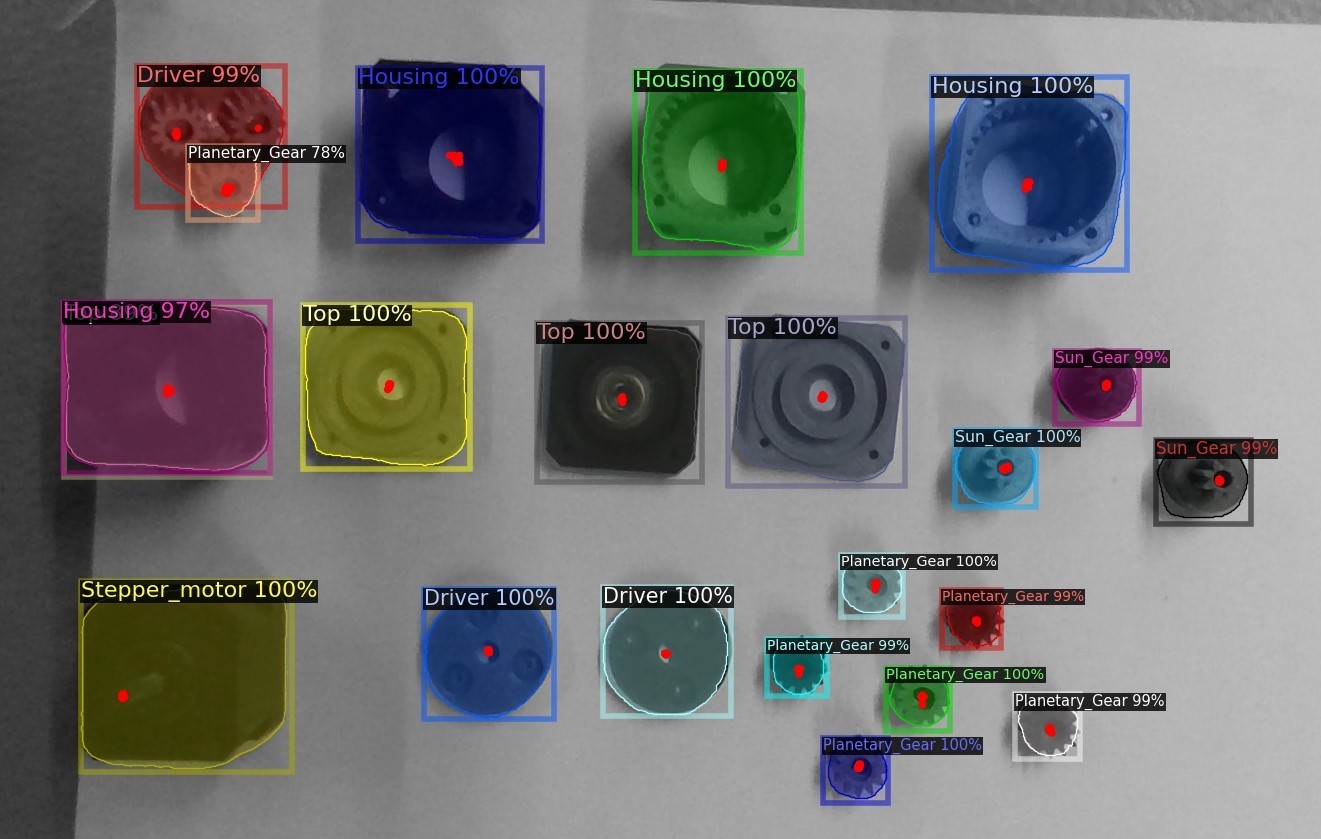} 
}

\caption{Results for the Diesel engine (top row) and planetary gear (bottom row) datasets. (a) and (c) depict examples of image annotations with polygons and key points. (b) and (d) depict the results of the trained  multi-task detection models (bounding box and segmentation mask with class number and confidence score, and red key points). 
\label{fig:annotation_detection_results}
}
\end{figure*}

        Fig. \ref{fig:diesel_detection} and \ref{fig:plan_detections} present the output of the trained multi-task models on the two use cases with results for object detection and segmentation, and key point estimations. This demonstrates that a single multi-task model can utilize multiple data annotation formats as part of the dataset, train a model and achieve successful detection outputs. 
        On average, the detection confidence is over 90 percent. Furthermore, as main beneficial outcome Table \ref{tab:fps} presents that the achieved frame rate of the multi-task model is higher than running multiple models in parallel. For high-resolution camera input (1280$\times$720), the difference is almost double, i.e., 8.7 FPS versus 4.8 FPS, respectively. Likewise, the memory requirements for a multi-task model is less than running multiple single-task models in parallel (i.e., 500 MB vs 700 MB).

        \begin{table}[t]
\caption{Frame rate and memory usage comparison between single- and multi-task models.}
\centering
\begin{tabular}{|p{0.8in}|c|c|c|}
\hline
\textbf{Model} & \textbf{Output} & \begin{tabular}[c]{@{}c@{}}\textbf{Inference} \\\textbf{(FPS)}\end{tabular} & \begin{tabular}[c]{@{}c@{}}\textbf{Memory} \\\textbf{(MB)}\end{tabular} \\ \hline
Faster R-CNN    & BB, class        & 11.1 & $\sim350$\\ \hline
Mask R-CNN      & Mask, BB, class & 10.5 & $\sim350$\\ \hline
Keypoint R-CNN  & Kps, BB, class  & 9.4  & $\sim350$\\ \hline
Multi-task model & Mask, kps, BB, class & 8.7 &  $\sim500$\\ \hline
\begin{tabular}[l]{@{}l@{}}Mask + keypoint \\(in parallel)\end{tabular} & Mask, kps, BB, class & 4.8 & $\sim700$\\ \hline
\end{tabular}
\label{tab:fps}
\end{table}

\subsection{Object Configuration and Depth Estimation}
        Two further functionalities are developed to demonstrate the use of multiple detection format outputs from the multi-task model. These are object configuration and depth estimation, based on the object segmentation masks and key points, respectively. 
        Segmentation masks generated by the multi-task model can be used to determine rocker arm orientations with respect to the image, with the 2nd order moment of the mask. The three estimated key points on the rocker arm then verify correct configuration of the rocker arm object with respect to the engine. The process for finding the outer side for the correct configuration is to minimize the axis of the smallest 2nd order moment, then apply a cross-product between the axis coordinates and predicted key points. The one key point with an opposite sign to the others indicates the outer side. Fig. ~\ref{fig:motivation} illustrates the correct orientation of the rocker arm on the engine and verifies (in)correct placement. 
        

        To estimate the depth of the rocker arm target, with respect to the camera, two methods were considered:  by segmentation mask or by key points (see Fig. \ref{fig:diesel_detection}). In both cases the depth value from corresponding RGB pixel coordinates are retrieved from a depth camera (Intel Realsense D435). For a segmentation mask, the entire mask area as depth range is taken and an average depth value is returned, and for a key point, a single depth value is returned. The key point approach provides 10 times less difference in depth range (i.e., $468 \pm 5mm$) as compared to the segmentation mask, and is therefore found to be more reliable. 

\subsection{Discussion}
The proposed approach enables images to be annotated by multiple different labels, including bounding boxes, polygons and key points, during a single annotation step, and store the result in a single convenient format (COCO). The annotations are then all included in the same data augmentation process to collect a dataset utilized as input for training a multi-task model. In effect, this provides a single model with multiple detection outputs in different detection formats (bounding box, segmentation masks and key points). Compared to single-task models that only consider single annotation format input and detection output, the following benefits are identified. 1. Time-saving for image annotation and augmentation. All annotations can be done in a single image, without changing any formats for different annotation types. For example, in case of a single task-model, one image would need to be annotated three times separately for three different annotation types. 2. Memory space requirements saving in model training. As only a single multi-task model needs to be trained instead of multiple single-task models, size of the model is reduced. 3. Combining detection outputs. Detection modalities from the multi-task model can be combined to solve additional detection tasks, not directly provided by the model. For example, depth and object configuration from key points as shown by our use cases, or other post-prediction applications such as boundary or grasp pose detection, or the estimation of object properties.

A current limitation of the annotation pipeline is that it relies on existing annotation (Label Studio \cite{LabelStudio}) and augmentation (Albumentations \cite{Albumentations}) tools. While these tools can be used under an open source license, access and support might change over time. Another limitation is that, still, annotation is a time-consuming step in the process of generating a learning-based detection model. Selection of which data to annotate and in which format (e.g., depth from pixel masks or key points) requires manual decisions and an iterative process to select a best outcome for the tasks. Automated and interactive annotation (e.g., based on machine learning \cite{cheng2018survey, Bhagat20181}) and semi-automated, interactive image annotation for visual detection exist \cite{Bragantini2022}, but are still in early stages of research. This is a natural next-step in developing assistive tools for model training and planned as future work.

\section{CONCLUSION}\label{conclusion}
Learning-based multi-task models require suitable tools for the collection and annotation of data, especially when data needs multiple labels. 
In this work we proposed a single framework that collects and annotates all required labels (bounding boxes, polygons and key points) from a single image and stores it as a single benchmark format (COCO). Augmentation can then utilize this single format to generate a full dataset. The approach is validated with two use cases, where three different annotation types are needed per image. Results show that multi-task models provide benefits with respect to inference time and memory usage as compared to multiple single-task models. 
In addition, results show that multi-task models allow to combine detection outputs and solve additional detection tasks from a single model, such as the estimation of depth and object configuration.

\section*{ACKNOWLEDGMENT}
Project funding was received from European Union's Horizon 2020 research and innovation programme, grant no. 871449 (OpenDR) and 871252 (METRICS).








\bibliographystyle{ieeetr}
\bibliography{references}

\end{document}